\documentclass{article}

    \PassOptionsToPackage{numbers, compress}{natbib}

\usepackage[preprint]{neurips_2025}

\usepackage[utf8]{inputenc} 
\usepackage[T1]{fontenc}    
\usepackage{hyperref}       
\usepackage{url}            
\usepackage{booktabs}       
\usepackage{amsfonts}       
\usepackage{nicefrac}       
\usepackage{microtype}      
\usepackage{xcolor}         
\usepackage{bm}
\usepackage{graphicx}
\usepackage{caption}      
\usepackage{amsmath}
\usepackage{wrapfig}
\usepackage{enumitem}
\usepackage{multirow}
\title{GenHOI: Generalizing Text-driven 4D Human-Object Interaction Synthesis for Unseen Objects}

\author{%
  Shujia Li\textsuperscript{1,2} \;\;\;\;\;\;\;\;\;\;\;\;
  Haiyu Zhang\textsuperscript{2,3} \;\;\;\;\;\;\;\; \;\;\;\;
  \textbf{Xinyuan Chen}\textsuperscript{2} \\
  \textbf{Yaohui Wang}\textsuperscript{2}\footnotemark[1]\;\;\;\;\;\;\;\;
  \textbf{Yutong Ban}\textsuperscript{1}\footnotemark[1]
 \\
    {\textsuperscript{1}Shanghai Jiao Tong University
    \textsuperscript{2}Shanghai AI Laboratory \ \ 
    \textsuperscript{3}Beihang University
    }\\
 \small\url{https://etach-qs.github.io/GenHOI_project/}
}
\begin{document}

\renewcommand{\thefootnote}{\fnsymbol{footnote}}
\footnotetext[1]{Corresponding author}

\maketitle

\begin{figure}[th]
\vspace{-1.5em}
\centering
\includegraphics[width=\linewidth, trim={0 0.5cm 0 0.5cm}, keepaspectratio]{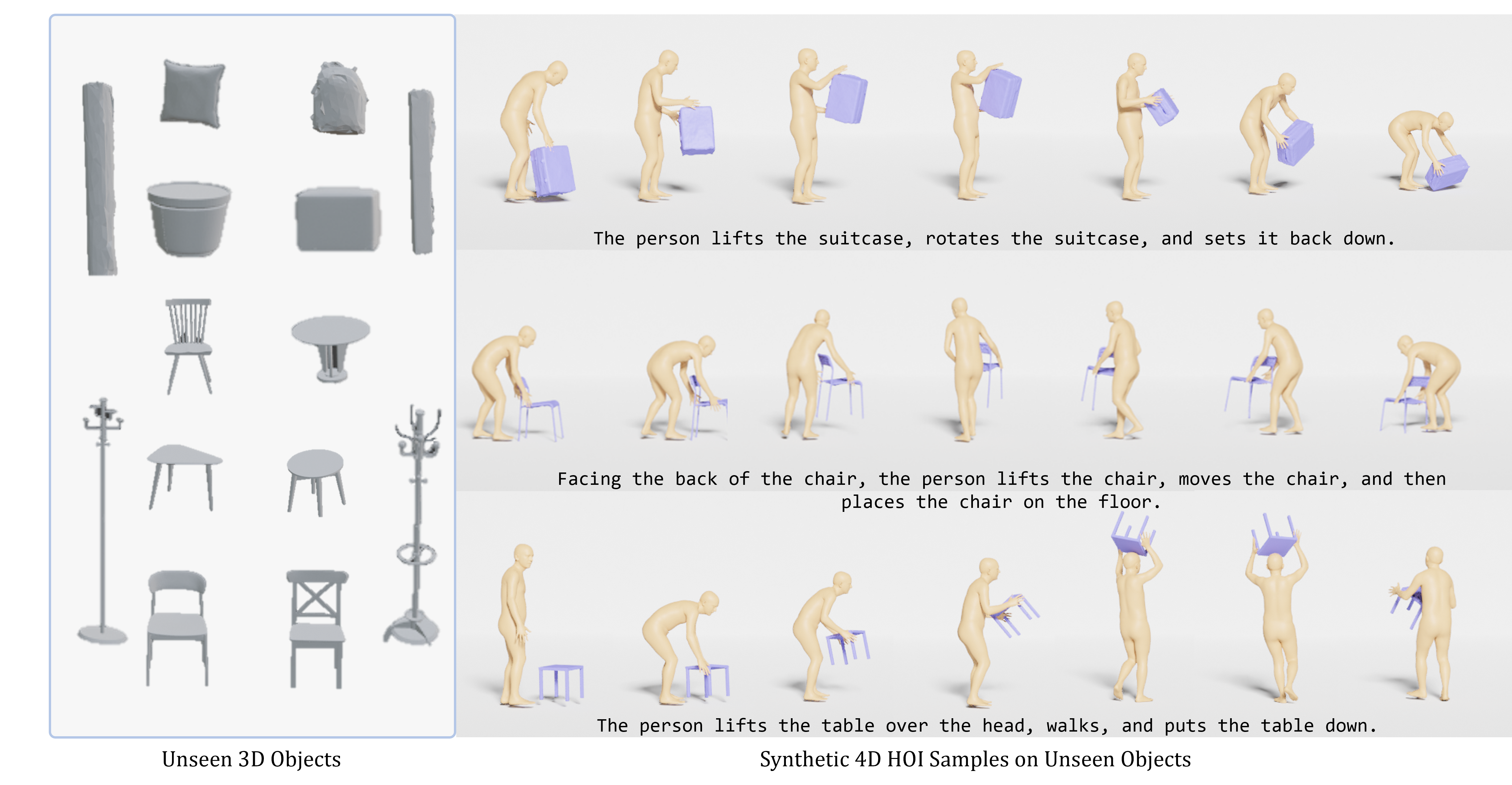}
\caption{
GenHOI synthesizes interaction between human and unseen object conditioned on text.
\label{fig:teaser}}
\end{figure}
\begin{abstract}
While diffusion models and large-scale motion datasets have advanced text-driven human motion synthesis, extending these advances to 4D human-object interaction (HOI) remains challenging, mainly due to the limited availability of large-scale 4D HOI datasets. In our study, we introduce GenHOI, a novel two-stage framework aimed at achieving two key objectives: 1) generalization to unseen objects and 2) the synthesis of high-fidelity 4D HOI sequences.
In the initial stage of our framework, we employ an Object-AnchorNet to reconstruct sparse 3D HOI keyframes for unseen objects, learning solely from 3D HOI datasets, thereby mitigating the dependence on large-scale 4D HOI datasets. Subsequently, we introduce a Contact-Aware Diffusion Model (ContactDM) in the second stage to seamlessly interpolate sparse 3D HOI keyframes into densely temporally coherent 4D HOI sequences. To enhance the quality of generated 4D HOI sequences, we propose a novel Contact-Aware Encoder within ContactDM to extract human-object contact patterns and a novel Contact-Aware HOI Attention to effectively integrate the contact signals into diffusion models. Experimental results show that we achieve state-of-the-art results on the publicly available OMOMO and 3D-FUTURE datasets, demonstrating strong generalization abilities to unseen objects, while enabling high-fidelity 4D HOI generation.

\end{abstract}

\section{Introduction}
\label{sec:intro}

Text-driven 4D HOI generation aims to synthesize realistic human motion and consistent object trajectories. This capability is critical for advancing applications in augmented and virtual reality (AR/VR), game development, and robotics. 
Recent advances in diffusion models and large-scale human motion datasets have led to significant progress in text-to-human motion generation. Building on these successes, pioneering studies \cite{li2023object, li2024controllable,song2024hoianimator,wu2024thor,peng2023hoi} on 4D HOI generation have attempted to replicate these achievements using diffusion models with 4D HOI datasets. However, these methods exhibit poor generalization ability to unseen objects due to the restricted object categories and the limited interaction patterns within the available 4D HOI datasets.

Specifically, BEHAVE \cite{bhatnagar2022behave}, CHAIRS \cite{jiang2023full}, and ARCTIC \cite{fan2023arctic} datasets predominantly contain 4D HOI sequences with rigid or articulated interactions and category-specific objects (e.g., chairs, boxes), while neglecting deformable or irregularly shaped objects that require distinct motion patterns and contact dynamics. GRAB \cite{taheri2020grab} only includes grasping-type interactions, overlooking the complex forms of actions in daily life. This narrow range of object categories and actions directly impacts model generalization. Models trained on these datasets \cite{peng2023hoi,taheri2020grab,wu2024thor,zhang2022couch,xu2023interdiff, cha2024text2hoi} can only generate 4D HOI sequences for specific objects and interactions, but struggle to adapt to unseen objects.

To reduce the reliance on large-scale 4D HOI datasets that encompass diverse object categories and interaction patterns, we propose a novel \textbf{Gen}eralized 4D \textbf{HOI} synthesis (\textbf{GenHOI}) framework. Specifically, we decompose the 4D HOI synthesis into two manageable sub-tasks: (1) recovering 3D HOI keyframes for unseen objects; (2) interpolating sparse 3D HOI keyframes into temporally coherent dense 4D HOI sequences. To deal with the two subtasks, we present a two-stage pipeline. In the first stage, we propose an Object-AnchorNet to learn human-object interaction patterns, which enables recovering 3D HOI keyframes when given human point clouds and object geometries. This network is trained on 3D HOI datasets, eliminating the need for large-scale 4D HOI datasets. In the second stage, we present a Contact-Aware diffusion model to interpolate the generated sparse 3D HOI keyframes into temporally coherent 4D HOI sequences. We propose a Contact-Aware Encoder to extract features from sparse 3D HOI keyframes, which implies human-object contact information, thereby facilitating the learning of the diffusion model. The contact information is efficiently integrated into diffusion model via a Contact-Aware HOI Attention. By utilizing the large-scale 3D HOI datasets and decouping the spatial and temporal modeling, we effectively reduce the reliance on large-scale 4D HOI dataset and achieve high-fidelity 4D HOI synthesis for unseen objects as shown in Fig. \ref{fig:teaser}. Our contributions can be summarized as follows:

\begin{itemize}[leftmargin=*,itemsep=0pt] 
    \item \textbf{A Novel Framework for Generalized 4D HOI Synthesis:} We present a two-stage pipeline that decouples spatial and temporal modeling, which enables natural and realistic 4D HOI synthesis for unseen objects.
    \item \textbf{3D HOI Keyframe Recovery:} We propose an Object-AnchorNet to recover 3D HOI keyframes from human point clouds and object geometries, which effectively reduces the reliance on large-scale 4D HOI datasets.
    \item \textbf{4D HOI Generation}: We propose a Contact-Aware Diffusion Model that interpolates sparse 3D HOI keyframes into temporally coherent dense 4D HOI sequences by utilizing contact patterns extracted from 3D HOI keyframes by a Contact-Aware Encoder and integrated by a Contact-Aware Attention.
    \item  Experimental results demonstrate that our method generates realistic 4D HOI sequences and achieves robust generalization to diverse unseen objects.
\end{itemize}


\section{Related Work}
\label{sec:related}

\subsection{Contextual Interaction Synthesis}
Recent research on human-object interaction (HOI) synthesis can be categorized into interactions with static and dynamic objects.

\noindent \textbf{Interactions with Static Objects.}
Synthesizing human motion in static 3D scenes has been widely studied using regression models, diffusion models, and reinforcement learning \cite{hassan2019resolving, hassan2021stochastic, huang2023diffusion, wang2022humanise,yi2024generating,zhang2022couch}. These methods generate interactions like sitting, lying, and navigating confined spaces but struggle with dynamic objects and adaptive environments. They cannot model scenarios where objects move, deform, or change states, such as pushing doors or rearranging furniture. While recent efforts have begun integrating dynamic object interactions \cite{peng2023hoi,li2023object,diller2024cg,geng2025auto}, real-time adaptability remains a challenge, limiting their ability to generalize in complex environments.  Addressing these gaps is essential for more realistic and versatile human-scene interaction synthesis.

\noindent \textbf{Interactions with Dynamic Objects.}
Synthesizing human-object interactions with dynamic objects has gained increasing attention \cite{xu2025intermimic,sui2025survey,zeng2025chainhoi,yang2024fhoifinegrainedsemanticaligned3d, deng2025human,cong2025semgeomo}. Early works primarily focused on motion prediction, where future interactions are forecasted based on past motion cues \cite{xu2023interdiff, wang2024move}. For instance, InterDiff models whole-body human-object interactions assuming deterministic interaction types given past motions. In contrast, OMOMO \cite{li2023object} introduces a conditional diffusion framework that generates human motion from object trajectories, yet it requires predefined object motion as input. Recent advancements have explored controllable synthesis of dynamic human-object interactions driven by textual inputs\cite{ghosh2023imos,li2024controllable,diller2024cg,lv2024himo,song2024hoianimator,peng2023hoi,wu2024human,liu2024core4d,liang2024intergen,xu2024inter,guo2024crowdmogen}. IMoS \cite{ghosh2023imos} generates full-body human and object motions from text, primarily focused on grasping. CHOIS \cite{li2024controllable} enhances control by synthesizing text-driven 4D interactions with sparse waypoints. More recent studies integrate physics-based and kinematic-based approaches \cite{Hodgins:2017:DOE, 2018-TOG-deepMimic, peng2021amp, wang2023physhoi, NEURIPS2024_918b9487} to model realistic whole-body interactions with dynamic objects. However, generating high-quality 4D HOI remains a challenging task due to the shortage of large-scale 4D HOI datasets and the inherent difficulty of ensuring physical plausibility.


\subsection{Zero-Shot Interaction Synthesis}

A major challenge in 4D HOI synthesis is the scarcity of large-scale 4D HOI annotated datasets. While datasets such as \cite{bhatnagar2022behave, fan2023arctic, huang2022intercap, jiang2023full,li2023object,zhang2024core4d,taheri2020grab} provide a foundation, they are significantly smaller than broader text-motion datasets \cite{guo2022generating,lin2023motion,mahmood2019amass}, limiting the capability of supervised learning to fully capture complex interactions. To address this, recent efforts explore zero-shot learning \cite{xu2024interdreamer, li2024zerohsi, kim2025david} to enhance motion generalization across diverse interactions. InterDreamer \cite{xu2024interdreamer} has recently proposed generating text-aligned HOI sequences in a zero-shot manner. However, the performance of their world model is constrained by the 4D HOI datasets used. Addtionally, recent state-of-the-art methods~\cite{li2024zerohsi, kim2025david} reconstruct human–object interactions by leveraging advanced video generative models to render human and object positions from video data, refining them frame by frame to produce coherent motion sequences. While effective, these approaches involve highly complex and time-consuming pipelines. In contrast, our method adopts a decoupled spatial–temporal modeling framework, which enables the efficient generation of semantically consistent 4D HOI sequences for unseen objects.

\section{Method}
\label{sec:method}

\begin{figure*}[t!]
\begin{center}
\includegraphics[width=\textwidth]{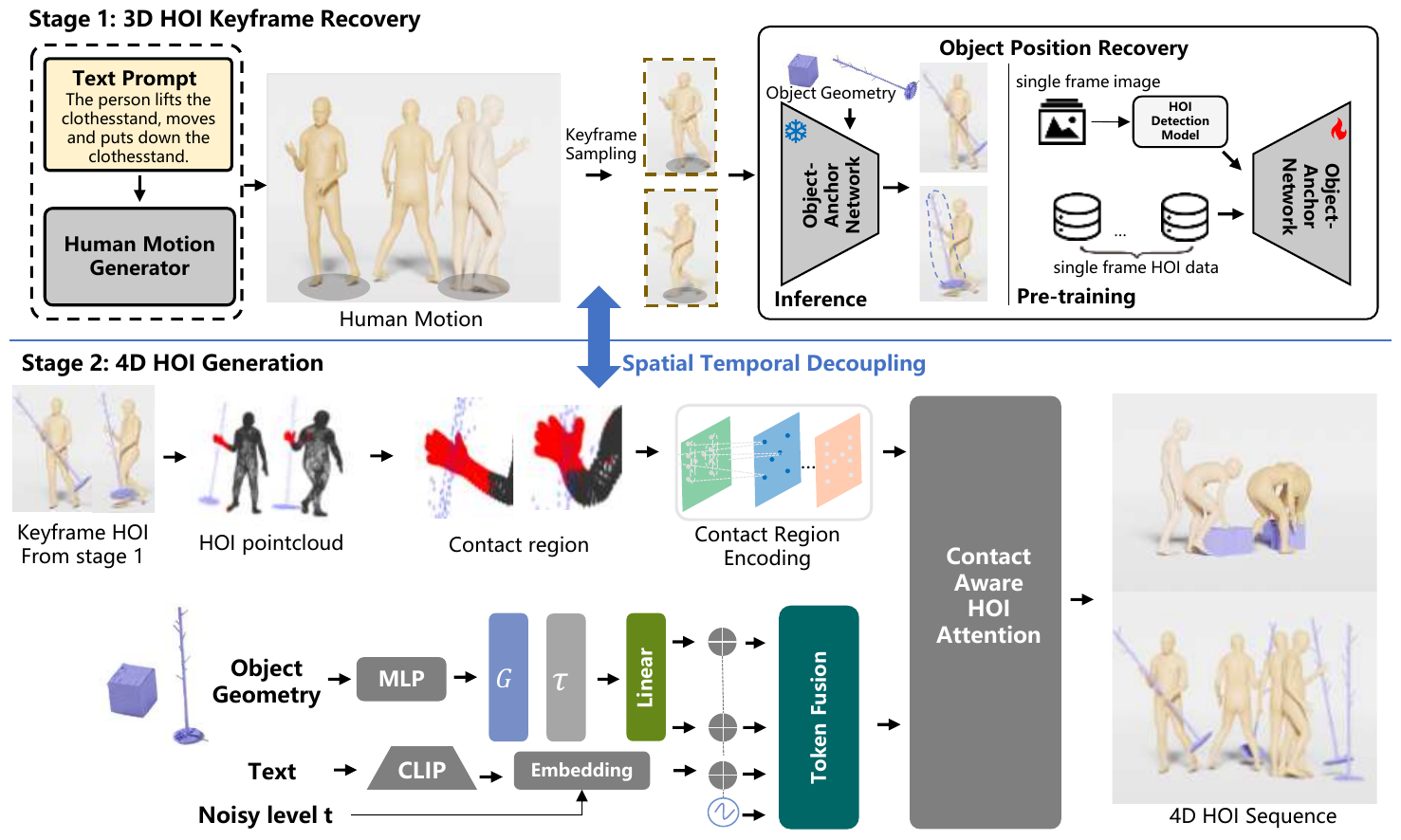}
\end{center}
\vspace{-2mm}
\caption{\textbf{GenHOI Overview.} In Stage 1, the method recovers keyframe HOI using object geometry and human pose priors. In Stage 2, the Contact-Aware Diffusion Model (ContactDM) synthesizes the 4D HOI sequence leveraging the keyframe HOI and the encoded contact embeddings. After training, the model can generalize to unseen objects given the object geometry and the associated text prompt.}
\label{fig:overview}
\end{figure*}

Our goal is to synthesise 4D human–object interaction (HOI) sequences conditioned on textual prompts and unseen objects. To this end, we propose a two-stage pipeline that decouples spatial and temporal modeling, thereby reducing dependence on large-scale 4D HOI datasets. In the first stage, we introduce Object-AnchorNet, which learns human–object interaction patterns and enables the recovery of 3D HOI keyframes with accurate spatial relationships by training on large-scale 3D HOI datasets (Sec.\ref{sec:3D_hoi}). Object-AnchorNet demonstrates strong generalizability to unseen objects, effectively reducing the learning burden of the second stage. In the second stage, we present a Contact-Aware Diffusion Model that interpolates the sparse 3D HOI keyframes into temporally coherent 4D HOI sequences. This is achieved by employing a Contact-Aware Encoder to encode the 3D HOI keyframes and a Contact-Aware HOI Attention to integrate these features into the ContactDM framework (Sec.\ref{sec:4D_hoi}). An overview of our method is shown in Fig.~\ref{fig:overview}.

\subsection{Preliminaries}

\textbf{4D Human and Object Interaction Representation.} We represent human representation as \( \boldsymbol{x}^{h} \in \mathbb{R}^{N \times D} \), where \( N \) is the number of frames and \( D \) is the dimensionality of human pose. At each frame \( n \), the human representation \( \boldsymbol{x}^h_n \) comprises global joint positions  \( \boldsymbol{j} \in \mathbb{R}^{N \times (J \times 3)} \) and local 6D continuous rotations \( \boldsymbol{q} \in \mathbb{R}^{N \times (J \times D_{\text{rot}})} \) based on the SMPL-X parametric model~\cite{zhou2019continuity}, which provides a compact and expressive representation of the human body. Object motion $\boldsymbol{x}^o_{n}$ is represented by its global 3D position (centroid) \( \boldsymbol{o} \in \mathbb{R}^{N \times 3} \) and rotation \( \boldsymbol{r} \in \mathbb{R}^{N \times D_{\text{rot}}} \). Formally, 4D HOI sequences are defined as:

\begin{equation}
\boldsymbol{x}^{h} = [\boldsymbol{j}, \boldsymbol{q}], \quad \boldsymbol{x}^{o} = [\boldsymbol{o}, \boldsymbol{r}],  \label{eq:1}
\end{equation}

\noindent where \( J \) is the number of human joints.

\subsection{3D HOI Keyframe Recovery}
\label{sec:3D_hoi}

\noindent\textbf{Human Motion Keyframes Sampling.} We first sample human motion sequences from generative models, existing human motion datasets, or real-world capture using motion sensors. Then, we try to sample keyframes by uniformly downsampling the obtained human motion sequences \(\boldsymbol{x}\). Specifically,  we select \(K=5\) keyframes through temporal averaging, aiming to preserve essential motion dynamics while reducing redundancy. This choice of \(K\) strikes a balance between computational efficiency and motion fidelity, as empirically validated in our experiments. For each keyframe, we reconstruct the human mesh using the SMPL-X model \cite{pavlakos2019expressive}, yielding $M_{1}$ spatial vertex positions denoted as $\boldsymbol{V}=\{ \boldsymbol{V}_i^h\in \mathbb{R}^{M_{1} \times 3} | i\in[1,...,K] \}$. By operating on sparse human motion keyframes rather than dense sequences, we reduce both error propagation and computational cost, while capturing a diverse range of interaction states.

\noindent\textbf{Object-AnchorNet.} Building on the sparse keyframes and their corresponding human point clouds, we propose Object-AnchorNet to recover object motion by learning the spatial relationships between human point clouds and object templates. Specifically, our Object-AnchorNet adapts Object Pop-up architecture \cite{petrov2023object} while removing the class label encoder to focus exclusively on geometric encoding of objects. It takes a human point cloud $\boldsymbol{V}_i^h$, an object template point cloud, and a text prompt \( \boldsymbol{c} \) as input, and predict the object’s pose to generate the target object point cloud $\boldsymbol{V}_i^o \in \mathbb{R}^{M_{2} \times 3}$, thereby forming a complete 3D HOI frame $(\boldsymbol{V}_i^h, \boldsymbol{V}_i^o)$. By training on point clouds that preserve detailed geometric information, rather than relying on coarse SMPL-X parameters and object poses, the network effectively captures fine-grained contact dynamics, which are critical for precise and realistic 3D HOI keyframe recovery. Additionally, Object-AnchorNet is trained on 3D HOI datasets composed of \textit{Grab}~\cite{taheri2020grab}, \textit{Behave}~\cite{bhatnagar2022behave}, and large-scale \textit{Open3DHOI} dataset~\cite{wen2025reconstructing}. This comprehensive collection spans 100+ daily object categories and diverse interaction types, providing rich coverage of human-object interaction scenarios. The multi-source training strategy significantly improves Object-AnchorNet’s generalizability to unseen object shapes and interaction patterns. By generating 3D HOI keyframes in the first stage, we reduce the learning burden in the second stage and enable to generated realistic 4D HOI sequences with unseen objects. For more details about Object-AnchorNet, please refer to the Appendix \ref{appendix}.

\subsection{4D HOI Generation}
\label{sec:4D_hoi}

In the second stage, we interpolate the sparse 3D HOI keyframes into temporally coherent motion sequences using a Contact-Aware Diffuion Model. Below, we detail the architecture of ContactDM, highlighting its two core components: a Contact-Aware Encoder and a Contact-Aware HOI Attention, both of which are critical for high-quality 4D HOI synthesis.

\noindent\textbf{Contact-Aware Diffusion Model.} The ContactDM $\mathcal{N}$ model follows the CHOIS architecture~\cite{li2024controllable} and is trained using the diffusion framework~\cite{ho2020denoising}. It generates temporally coherent 4D HOI sequences $(\boldsymbol{x}^{h}, \boldsymbol{x}^{o})$ and contact labels $\boldsymbol{H} \in \mathbb{R}^{N\times2}$ for the left and right hands, conditioned on 3D HOI keyframes $\mathcal{V}_{K}$, object geometry $G$, and textual prompts $c$. The object geometry utilizes the Basis Point Set (BPS) representation and is projected to low-dimensional vector via a multilayer perceptron (MLP). Textual prompts are encoded using CLIP~\cite{radford2021learning}, which extracts semantic information to guide generation. The projected object features and CLIP embeddings are fused with the 4D HOI sequences through a token fusion module that employs a self-attention mechanism. Additionally, we propose a Contact-Aware Encoder to extract fine-grained contact-aware features from sparse 3D HOI keyframes, capturing subtle patterns of human–object interaction. These features are then integrated using a Contact-Aware HOI Attention module via a cross-attention mechanism. This design ensures the precise incorporation of spatial and contact cues, enabling the generation of realistic and temporally coherent 4D HOI sequences. During inference, we leverage reconstruction guidance with contact labels to refine the 4D HOI sequences as detailed in ~\cite{li2024controllable}.

\noindent\textbf{Contact-Aware Encoder.} Although diffusion models based on pose-based representations are lightweight and computationally efficient, they struggle to capture fine-grained details of 3D HOI contact regions. To address this limitation, we propose a Contact-Aware Encoder that encodes HOI point clouds, extracting contact-aware features to enrich the representation with precise spatial and interaction information essential for realistic synthesis. Specifically, given the 3D HOI keyframe $(\boldsymbol{V}_{i}^h, \boldsymbol{V}_{i}^o)$, we use PointNet++ \cite{qi2017pointnet++} to encode the 3D HOI point clouds. However, directly encoding the human and object point clouds using PointNet++, resulting in significant memory overhead. To mitigate this, we adopt an efficient sampling strategy. First, we downsample the object point cloud $\boldsymbol{V}^o_i \in \mathbb{R}^{M_2 \times 3}$ to $\boldsymbol{\hat{V}}_i^{o} \in \mathbb{R}^{M_o \times 3}$ by selecting the $M_o$ farthest points, while preserving the object's geometric shape. Second, to accurately infer contact relationships, we employ the K-Nearest Neighbors (KNN) algorithm to sample $M_h$ points from the human point cloud, that are closest to the object point cloud. The resulting subset of human points is denoted as $\boldsymbol{\hat{V}}^{h}_{i} \in \mathbb{R}^{M_h \times 3} = \text{KNN}(\boldsymbol{V}^h_{i}, \boldsymbol{V}^o_{i})$. In our experiment, $M_o = 500, M_h = 1000$. To distinguish human and object vertices, we introduce one-hot encoding:
\begin{equation}
    \begin{split}
        \tilde{\boldsymbol{V}}_i^h &= \text{concat}(\boldsymbol{\hat{V}}^{h}_{i}, \boldsymbol{1}_{M_h}) \in \mathbb{R}^{M_h \times 4}, \\
        \tilde{\boldsymbol{V}}_i^o &= \text{concat}(\boldsymbol{\hat{V}}^{o}_{i}, \boldsymbol{0}_{M_o}) \in \mathbb{R}^{M_o \times 4}
    \end{split}
\end{equation}
where $\boldsymbol{1}_{M_h}$ and $\boldsymbol{0}_{M_o}$ denote vectors of ones and zeros, respectively, for human and object points. The final input point cloud is written as:
\begin{equation}
    \tilde{\boldsymbol{V}}_i = \text{concat}(\tilde{\boldsymbol{V}}_i^h, \tilde{\boldsymbol{V}}_i^o) \in \mathbb{R}^{(M_h+M_o) \times 4}
\end{equation}

The obtained point cloud is then processed by a point cloud encoder~\cite{qi2017pointnet++}, employing a multi-scale grouping strategy to extract hierarchical spatial features:
\begin{equation}
    \boldsymbol{F}_i = \text{PointEncoder}(\tilde{\boldsymbol{V}}_i) \in \mathbb{R}^{d}
\end{equation}
where $\boldsymbol{F}_i$ represents the encoded per-frame feature, and $d$ is the output feature dimension. The point cloud encoder aggregates features at multiple scales using local neighborhoods, ensuring both geometric and contact-aware representation learning.

\noindent\textbf{Contact-Aware HOI Attention.} The Contact-Aware HOI Attention is a module bridges the encoded contact-aware features $\boldsymbol{F} = \text{concat}(\boldsymbol{F}_i | i=1, \dots, K)$ and the conditional diffusion model through a cross-attention mechanism. In contrast to addictive embedding fusion method that merely adds conditioning embeddings to the input in a static manner, the cross-attention mechanism facilitates dynamic alignment between feature representations $\boldsymbol{F}$ and the latent variables of the diffusion model. This approach enables more precise feature integration, as demonstrated in the supplementary materials \ref{appendix}. Specifically, we project $\boldsymbol{F}$ as keys $\boldsymbol{K}$ and values $\boldsymbol{V}$, while the pose embeddings $\boldsymbol{E}_{\text{pose}}$ is projected as queries $\boldsymbol{Q}$. This allows the diffusion model to selectively attend to critical contact regions during generation, ensuring that fine-grained spatial information guides the synthesis process. The fused feature $\boldsymbol{F}_{\text{fused}}$ is computed as:

\begin{equation}
\boldsymbol{F}_{\text{fused}} = \text{Softmax}\left(\frac{\boldsymbol{Q}\boldsymbol{K}^T}{\sqrt{d}}\right)\boldsymbol{V}
\end{equation}

\noindent where $\boldsymbol{Q}$, $\boldsymbol{K}$, and $\boldsymbol{V}$ are linear projections of $\boldsymbol{E}_{\text{pose}}$, $\boldsymbol{F}$, and $\boldsymbol{F}$, respectively.

\section{Experiment}
\label{sec:exp}

In this section, we first present the implementation details of our method. In addition, we demonstrate the effectiveness and generalizability of our approach through quantitative comparisons and user studies. We also conduct ablation studies to evaluate the different components of our method.

\subsection{Implementation Details}
\noindent \textbf{Evaluation Metrics.} We evaluate 4D HOI sequences from multiple perspectives. Specifically, we assess the quality of human motion using the Foot Sliding Score (FS), Fréchet Inception Distance (FID), and R-precision (Rprec). To measure interaction accuracy, we use precision ($C_{\text{prec}}$), recall ($C_{\text{rec}}$), F1 score ($C_{F_1}$), the contact percentage ($C\%$), and hand-object penetration score ($P_{\text{hand}}$). To quantify the deviation from ground truth, we report the mean per-joint position error (MPJPE), root joint translation error (\(T_{\text{root}}\)), object position error (\(T_{\text{obj}}\)), and object orientation error (\(O_{\text{obj}}\)).

\noindent\textbf{Baselines.} We adapt InterDiff~\cite{xu2023interdiff}, MDM~\cite{tevet2023human}, HOI-Diff~\cite{peng2023hoi}, OMOMO~\cite{li2023object}, and CHOIS~\cite{li2024controllable} as baselines for comparison. InterDiff predicts 4D HOI sequences based on past frames. MDM only generates human motion. OMOMO synthesizes 4D HOI conditioned on predefined object trajectories. HOI-Diff incorporates an affordance-aware module to improve contact precision. CHOIS produces 4D HOI sequences given the initial human pose and object waypoints. To ensure fair evaluation, we modify InterDiff to support text conditioning, extend MDM to HOI data by incorporating our object geometry representation.

\noindent \textbf{Datasets.} In the first stage, we train Object-AnchorNet on a mixture of 3D HOI datasets, including Behave \cite{bhatnagar2022behave}, Grab \cite{taheri2020grab}, and Open3DHOI \cite{wen2025reconstructing}. These datasets encompass a broad range of object categories with varying scales, as well as diverse interaction types, including carrying, lifting, sitting, and playing. In the second stage, we train the Contact-aware Diffusion Model on 4D HOI sequences from OMOMO \cite{li2023object} to learn fundamental action dynamics.

\noindent \textbf{Training Settings.} Our Contact-Aware Diffusion Model is built upon the diffusion transformer architecture~\cite{ho2020denoising} comprising 4 transformer blocks with a latent dimension of 1024. The Contact-Aware HOI Attention module employs a multi-head attention mechanism with 4 heads, each having a dimensionality of 256, resulting in a total attention dimension of 1024. We use the Adam optimizer with a learning rate of $1\times10^{-4}$ to train the model. Training is conducted on a single NVIDIA RTX 4090 GPU with a batch size of 32 for a total of 50,000 steps.

\begin{table*}[t!]
 \caption{\textbf{Interation synthesis} on the seen objects. OMOMO-GT utilizes
ground truth object motion as input for OMOMO.} 
    \label{tab:single_window_cmp_seen}
    \centering 
\footnotesize{
\setlength{\tabcolsep}{1pt}
  \resizebox{.9\textwidth}{!}{ 
\begin{tabular}{lcccccccccccc} 
 \toprule 
 & \multicolumn{3}{c}{Human Motion} & \multicolumn{5}{c}{Interaction} & \multicolumn{4}{c}{GT Difference}  \\
 \cmidrule(lr){2-4}\cmidrule(lr){5-9}\cmidrule(lr){10-13}
 Method  & FS$\downarrow$  & $R_{prec}\uparrow$ & $FID\downarrow$ & $C_{prec}\uparrow$ & $C_{rec}\uparrow$ & $C_{F_1}\uparrow$ & $C_{\%}$ & $P_{hand}\downarrow$ & MPJPE$\downarrow$  & $T_{root}\downarrow$  & $T_{obj}\downarrow$ & $O_{obj}\downarrow$   \\
        \midrule
         Interdiff~\cite{xu2023interdiff}  & 0.42 & 0.08 & 208.0 & 0.63 & 0.28 & 0.33 & 0.27 & 0.55 & 25.91 & 63.44 & 88.35 & 1.65   \\
        MDM~\cite{tevet2023human}   & 0.48 & 0.51 & 6.16 & 0.72 & 0.47 & 0.53 & 0.43 & 0.66 & 17.86 & 34.16 & 24.46 & 1.85   \\
        HOI-Diff~\cite{peng2023hoi} & 0.37 & 0.52 & 2.14 & 0.77 & 0.69 & 0.64 & 0.49 & 0.69 & 17.44 & 32.28 & 22.75 & 1.51 \\
        OMOMO-GT~\cite{li2023object}  & 0.40 & 0.54 & 4.19 & 0.77 & 0.66 & 0.67 & 0.59 & 0.55 & 15.82 & 24.75 & 0.0 & 0.0  \\
        CHOIS~\cite{li2024controllable} & \textbf{0.35} & \textbf{0.64} & 0.69 & 0.80 & 0.64 & 0.67 & 0.54 & 0.59 & 15.30 &  24.43 & 12.53 & 0.99 \\
        \midrule
        GenHOI w/o CA& 0.39 & 0.58 & 0.76 & 0.77 & 0.75  & 0.63 & 0.61 & 0.56 & 17.33 &  25.14 & 17.85 & 1.03 \\
        GenHOI (Full)& 0.38 & 0.61 & \textbf{0.41} & \textbf{0.82} & \textbf{0.75} & \textbf{0.77} & \textbf{0.64} & 0.66 & \textbf{10.64} &  \textbf{11.62} & \textbf{6.39} & \textbf{0.81} \\
        \bottomrule
\end{tabular}
}
}
\end{table*}

\begin{table*}[t!]
\small
\caption{\textbf{Interation synthesis} on the unseen objects of the OMOMO dataset. The best performance is indicated in bold, with the second-best performance being underlined.} 
    \label{tab:single_window_cmp_unseen}
    \centering 
\footnotesize{
\setlength{\tabcolsep}{1pt}
  \resizebox{.9\textwidth}{!}{ 
\begin{tabular}{lcccccccccccc} 
 \toprule 
 & \multicolumn{3}{c}{Human Motion} & \multicolumn{5}{c}{Interaction} & \multicolumn{4}{c}{GT Difference}  \\
 \cmidrule(lr){2-4}\cmidrule(lr){5-9}\cmidrule(lr){10-13}
 Method  & FS$\downarrow$  & $R_{prec}\uparrow$ & $FID\downarrow$ & $C_{prec}\uparrow$ & $C_{rec}\uparrow$ & $C_{F_1}\uparrow$ & $C_{\%}$ & $P_{hand}\downarrow$ & MPJPE$\downarrow$  & $T_{root}\downarrow$  & $T_{obj}\downarrow$ & $O_{obj}\downarrow$   \\
        \midrule
        OMOMO-GT~\cite{li2023object} &  0.43 & 0.52 & 4.27 &  0.18 &  0.04 & 0.06  &  0.12 &  0.24 &  31.07 & 77.62  & 0.0  & 0.0 \\
        CHOIS~\cite{li2024controllable}& 0.50 & \textbf{0.61} & 0.74 & 0.76 & 0.59 & 0.64 & 0.56 & \textbf{0.13} & 16.50 &  28.78 & \textbf{14.29} & \textbf{1.04} \\
        \midrule
        GenHOI w/o CA& 0.51 &  0.60 & 0.77 & 0.72 & 0.59 & 0.63 & 0.57 & 0.16 & 16.62 & 29.33  & 15.01 & \underline{1.05} \\
        GenHOI-GT& \textbf{0.42} & 0.58 & \textbf{0.44} & \textbf{0.79} & \textbf{0.62} & \textbf{0.67} & \textbf{0.58} & 0.20 & \textbf{16.00} &  \textbf{26.75} & \underline{14.51} & 1.14 \\
        GenHOI (Full)&0.44 & 0.59 & \underline{0.52} & \textbf{0.79}& \underline{0.61}& \underline{0.66}& \textbf{0.58} & 0.18 & \underline{16.07} & \underline{26.92} & 14.98 & 1.20 \\
        \bottomrule
\end{tabular}
}
}
\end{table*}

\subsection{Comparisons on 4D HOI Generation.}

\noindent\textbf{Seen Objects.} We begin by comparing our method with other methods on the 4D HOI generation task on seen objects. In this experiment, we train and test on all 15 objects in the OMOMO dataset. Table \ref{tab:single_window_cmp_seen} presents the quantitative results. In terms of human motion, our method significantly outperforms MDM, which focuses on modeling human motion. Compared to CHOIS, our model generates more realistic human motion, as evidenced by improved Fréchet Inception Distance (FID) scores. Regarding human-object interactions, our method achieves the best results across multiple metrics, including $C_{\text{prec}}$, $C_{\text{rec}}$, $C_{F_1}$, and $C_\%$, indicating that our model effectively captures the intricate patterns of human-object interactions. Furthermore, metrics on GT Difference validate that our generated 4D HOI sequences closely align with ground truth. It demonstrates that our 4D HOI sequences are more realistic and natural.

\begin{wraptable}[13]{r}{0.6\linewidth}
\small
\caption{\textbf{Interaction Synthesis} on unseen objects of the 3D-FUTURE dataset~\cite{fu20213d}.In OMOMO* method, we generate evaluation conditions by integrating these objects with motion from the test set of the OMOMO dataset, following \cite{li2024controllable}. } 
\label{tab:3D-FUTURE}
\centering
\begin{tabular}{lcccc} 
\toprule 
Method & $C_{\%}$ & $P_{hand}\downarrow$ & MPJPE$\downarrow$ & $T_{root}\downarrow$ \\
\midrule
OMOMO*~\cite{li2023object}& 0.50 & 0.18 & 20.03 & 30.41 \\
CHOIS~\cite{li2024controllable} & 0.48 & 0.15 & 16.39 & 24.52 \\
\midrule
GenHOI w/o CA & 0.45 & 0.22 & 17.46 & 26.75 \\
GenHOI (Full) & \textbf{0.58} & \textbf{0.12} & \textbf{10.82} & \textbf{11.77} \\
\bottomrule
\end{tabular}
\end{wraptable}

\noindent\textbf{Unseen Objects.} We evaluate the generalization ability of our method on unseen objects using two test settings: (1) Cross-object generalization on the OMOMO dataset, where the models are trained on 10 objects and tested on the remaining 5 unseen objects. The results are reported in Table~\ref{tab:single_window_cmp_unseen}; (2) Cross-dataset generalization, where the models are trained on all 15 objects from OMOMO and tested on 17 selected objects from the 3D-FUTURE dataset. Since the 3D-FUTURE dataset contains object geometries without corresponding 4D HOI sequences, we can only evaluate on a subset of metrics: $C_\%$, $P_{\text{hand}}$, MPJPE, and $T_{\text{root}}$. The results are shown in Table~\ref{tab:3D-FUTURE}.

\begin{figure*}
\centering
\vspace{-4mm}
\includegraphics[width=\textwidth]{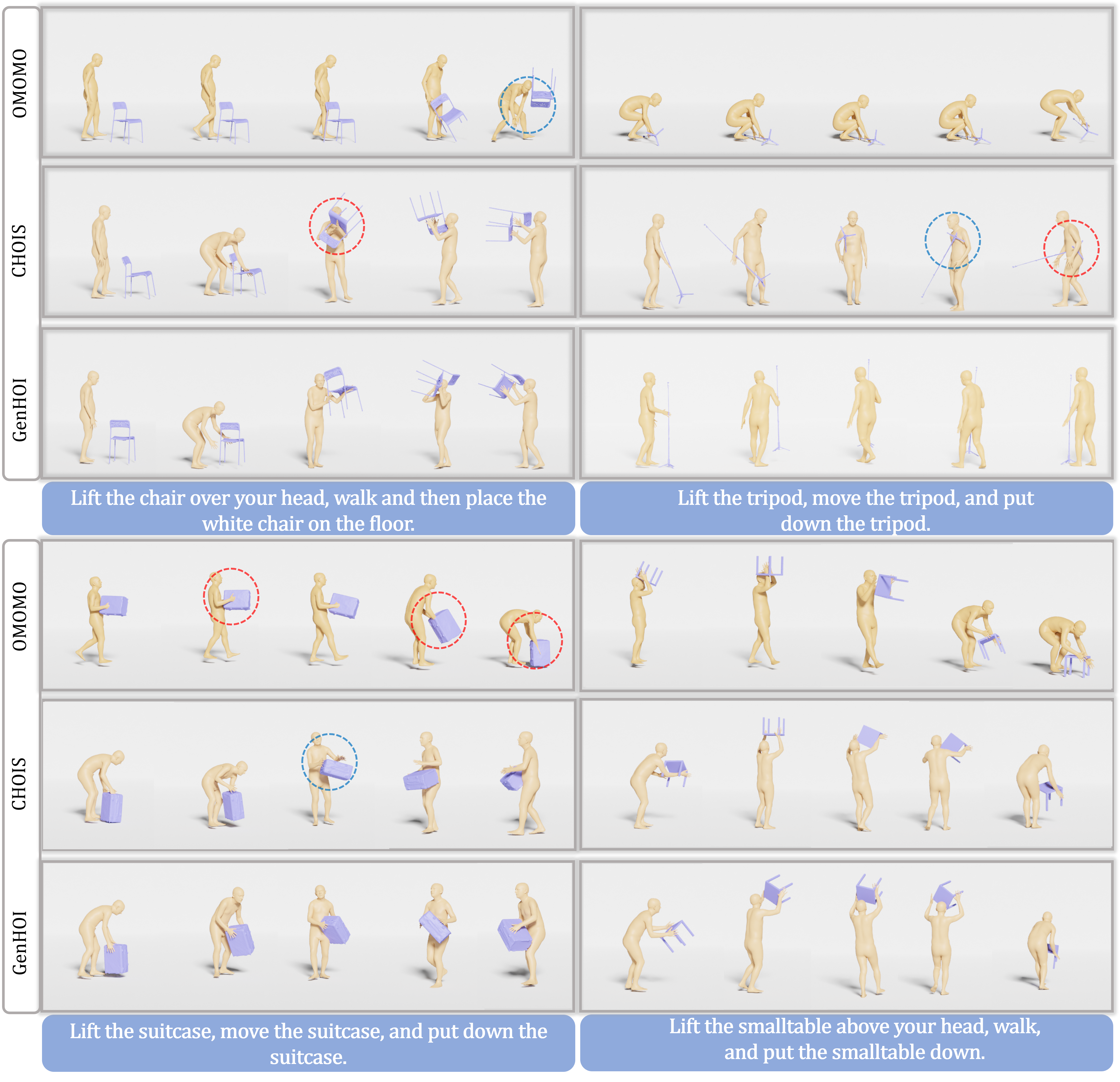}

\caption{Examples of synthetic motions for qualitative evaluation.}
\label{fig:comparisons}

\end{figure*}

As shown in Table~\ref{tab:single_window_cmp_unseen}, OMOMO exhibits significant object floating and penetration artifacts, resulting in substantially worse Interaction metrics compared to our method. This highlights its limited generalization capability. Under the more challenging cross-dataset setting (Table~\ref{tab:3D-FUTURE}), our method consistently outperforms all baselines across the four evaluation metrics, which assess both interaction quality and realism. These results underscore the strong generalization ability of our framework on unseen objects. We further provide qualitative comparisons with existing methods~\cite{li2023object,li2024controllable}. As illustrated in Fig.\ref{fig:comparisons}, OMOMO suffers from penetration issues with previously unseen objects such as suitcases and chairs, while CHOIS often displays clear hand-object separation and object floating artifacts—particularly visible with the triplot object in the second row of Fig.\ref{fig:comparisons}. In contrast, GenHOI generates physically plausible and semantically coherent 4D HOI sequences, demonstrating the effectiveness of our spatial-temporal decoupling strategy in modeling fine-grained human-object interaction dynamics and achieving strong generalizability to unseen scenarios.

\begin{figure*}[t]
\centering

\includegraphics[width=\textwidth]{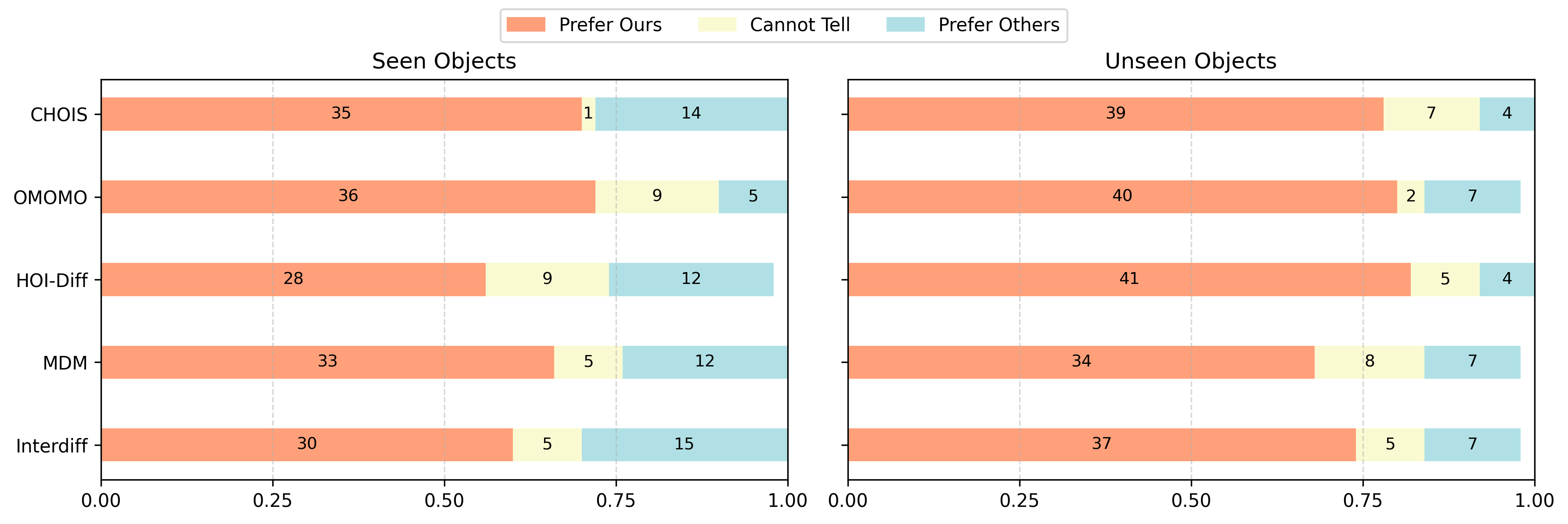}

\caption{Results of human perceptual studies. The numbers shown in the chart
represent the percentage (\%) over motion preferences.}
\label{fig:userstudy}
\vspace{-2mm}
\end{figure*}

\noindent\textbf{User Study} We also conduct a user study to further evaluate the qualitative preference for our method. Specifically, we sample 20 4D HOI sequences per method, covering 5 seen objects and 5 unseen objects. For each object, we generate 4D HOI sequences using 2 distinct text prompts. 50 participants are recruited to assess the quality of the generated sequences and indicate their preference among three choices. As shown in Figure~\ref{fig:userstudy}, the results clearly demonstrate that our model is consistently preferred over existing methods, validating the superiority of our approach in producing natural and coherent 4D HOI sequences.

\subsection{Evaluations on 3D HOI Keyframe Recovery}
\begin{figure}[h]
\begin{center}
\includegraphics[width=1.0\textwidth]{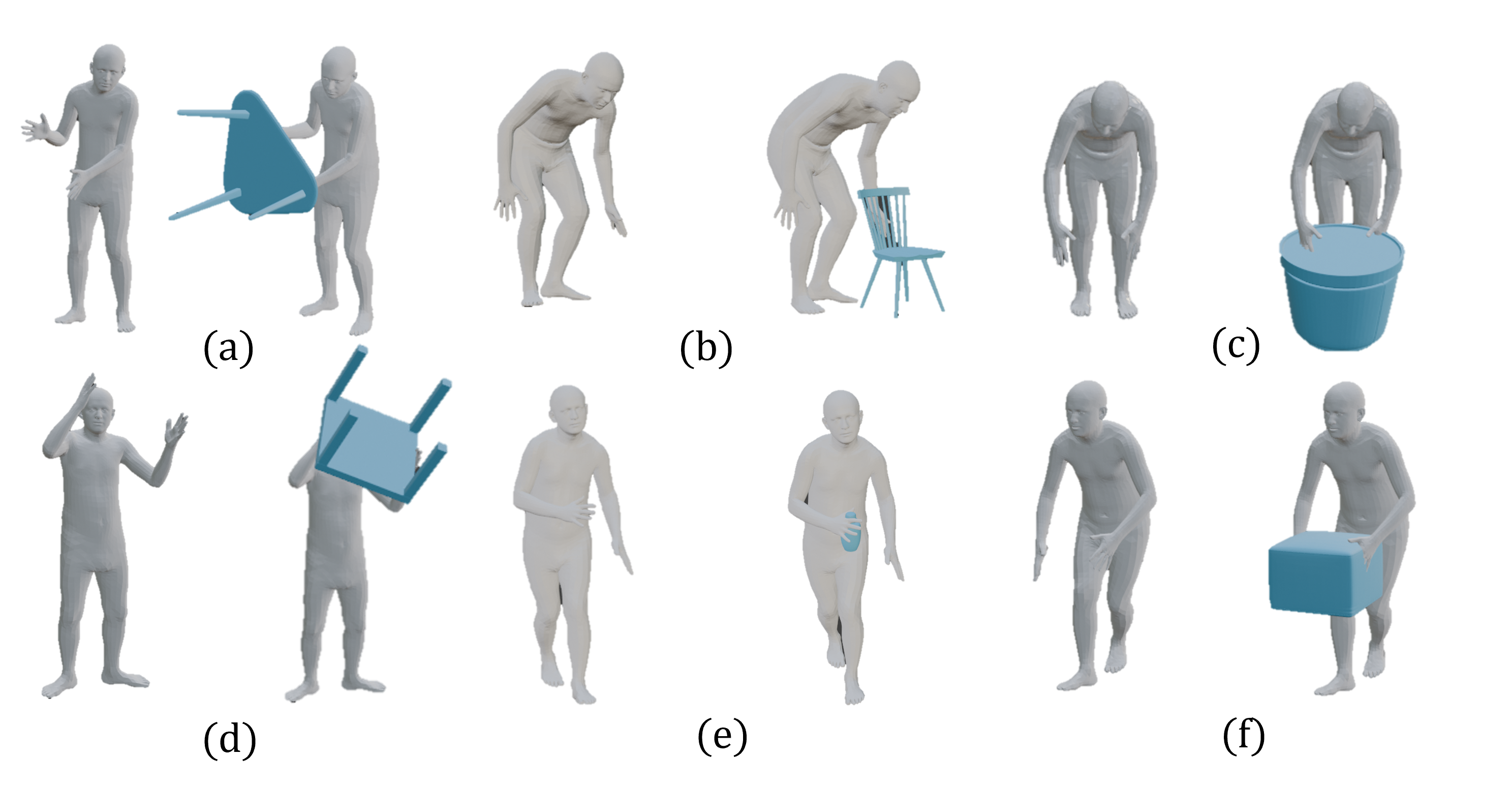}

\caption{Results of our Object-AnchorNet. Objects (a)-(e) are from 3D-FUTURE, and (f) is from OMOMO.}
\label{fig:objectanchor}
\end{center}
\vspace{-6mm}
\end{figure}

We evaluate the performance of Object-AnchorNet in recovering 3D HOI keyframes using unseen objects from the 3D-FUTURE dataset and single human frames sampled from the OMOMO dataset. As shown in Fig.~\ref{fig:objectanchor}, our method accurately estimates object positions and rotations, demonstrating strong generalization ability to unseen objects. The recovered object placements are natural and plausible, with no visible floating or penetration, such as correctly grasping a chair, lifting a bucket, or carrying a box. This demonstrates the model’s strong generalization to objects never seen during training. It is worth noting that this robust 3D HOI keyframe recovery forms a crucial foundation for the second stage, facilitating the generation of physically plausible and semantically coherent 4D HOI sequences on unseen objects.

\subsection{Ablation Studies}

GenHOI consists of several key designs: (1) a baseline model without the Contact-Aware Encoder and Contact-Aware HOI Attention (GenHOI w/o CA), (2) number of 3D HOI keyframes, (3) Contact-Aware Encoder, and (4) Contact-Aware HOI Attention. In the following, we conduct ablation studies to validate the contribution of each design.


\noindent\textbf{Baseline Model.} By omitting the 3D HOI keyframe recovery stage and removing both the Contact-Aware Encoder and Contact-Aware HOI Attention, our method (w/o CA) reduces to a diffusion model conditioned solely on object geometry and text. As shown in Table~\ref{tab:single_window_cmp_seen}, Table~\ref{tab:single_window_cmp_unseen}, and Table~\ref{tab:3D-FUTURE}, the absence of 3D HOI keyframes as a condition signal makes it difficult for the model to capture meaningful human-object interaction patterns, resulting in significantly worse performance on both seen and unseen objects. These results highlight the necessity of our decoupled spatial-temporal modeling framework.

\noindent\textbf{Number of 3D HOI Keyframes.}  We systematically evaluate the impact of the number of keyframe $K$ when interpolating the sparse 3D HOI keyframes to 120-frames 4D HOI sequences. As shown in Table~\ref{tab:different k}, when $K < 5$, these keyframes fail to sufficiently capture the full dynamics of human-object interactions, which increases the difficulty of interpolation in the second stage. As illustrated in Fig.~\ref{fig:teaser}, a single 3D HOI keyframe can only represent the interaction at a specific moment, which is insufficient to represent the subsequent interaction progression. In contrast, using a larger number of 3D HOI keyframes alleviates the learning burden of the diffusion model and results in improved synthesis quality and generalization ability. However, when $K > 5$, the prediction error of 3D HOI keyframes increases with the number of keyframes, which negatively impacts the final 4D HOI generation. The choice of $K=5$ offers an optimal trade-off, capturing sufficient interaction dynamics while maintaining prediction errors within a controllable range.

\noindent\textbf{Contact-Aware Encoder.} To validate the effectiveness of our Contact-Aware Encoder, we compared it with a variant that uniformly sampling points from 3D HOI keyframes. The results are shown in Table.~\ref{tab:condition}. We observed that under uniform sampling, the ContactDM struggles to converge. We hypothesize that this is due to the lack of emphasis on the nearest points between human and objects, which are critical for representing human-object contact. As a result, the encoder focuses more on global body posture rather than the fine-grained local interactions between specific body parts and the object,  leading to training difficulty of ContactDM.

\noindent \textbf{Contact-Aware HOI Attention.} We explore two strategies to integrate contact-aware features $\{\boldsymbol{F}_i\}_{i=1}^K$ into our diffusion model: \textit{additive embedding fusion} and \textit{cross-attention conditioning}, as shown in Table~\ref{tab:condition}. For the additive fusion strategy, we compute a global feature embedding $\bar{\boldsymbol{F}} = \frac{1}{T} \sum_{i=1}^T \boldsymbol{F}_i$, which is added to the diffusion timestep embedding and the text embedding. Although this strategy is computationally efficient, it struggles to inject contact-aware information accurately and may introduce interference among embeddings. In contrast, cross-attention conditioning aligns the 4D HOI latents with contact-aware features $\boldsymbol{F}$ through a cross-attention mechanism. This enables temporally and semantically precise conditioning, thereby enhancing the fidelity and realism of the generated 4D HOI sequences.

\begin{table*}[t!]
\small
\caption{Comparison of interaction quality for different number of keyframes on OMOMO dataset.} 
\label{tab:different k}
\centering 
\footnotesize{
\setlength{\tabcolsep}{2pt}
\resizebox{.9\textwidth}{!}{ 
\begin{tabular}{lcccccccccc} 
\toprule 
 & \multicolumn{5}{c}{Interaction} & \multicolumn{4}{c}{GT Difference}  \\
 \cmidrule(lr){2-6}\cmidrule(lr){7-10}
 Keyframe Numbers & $C_{prec}\uparrow$ & $C_{rec}\uparrow$ & $C_{F_1}\uparrow$ & $C_{\%}$ & $P_{hand}\downarrow$ & MPJPE$\downarrow$  & $T_{root}\downarrow$  & $T_{obj}\downarrow$ & $O_{obj}\downarrow$ \\
\midrule
K = 1 & 0.80 & 0.69 & 0.71 & 0.60 & 0.62 & 13.94 & 18.22 & 10.85 & 0.95\\
K = 3 & 0.81 &\textbf{ 0.75} & 0.76 & \textbf{0.64} & \textbf{0.63} & 11.40 & 12.59 & 7.68 & 0.82 \\
K = 5 & \textbf{0.82} & \textbf{0.75} & \textbf{0.77} & \textbf{0.64} & 0.66 & \textbf{10.64} & \textbf{11.62} & \textbf{6.39} & \textbf{0.81} \\
K = 8 & \textbf{0.82} & 0.67 & 0.71 & 0.57 & \textbf{0.63} & 10.71 & 11.66 & 9.89 & 0.87 \\
\bottomrule
\end{tabular}
}
}
\end{table*}

\begin{table*}[t!] 
\small
\caption{Comparison of Conditioning Strategies for 4D HOI Generation on OMOMO dataset.}  
\label{tab:condition}
\centering  
\footnotesize{
\setlength{\tabcolsep}{3pt}
\renewcommand{\arraystretch}{1.2}
\resizebox{.9\textwidth}{!}{
\begin{tabular}{llccccccccc}  
\toprule   

\multicolumn{2}{c}{} 
& \multicolumn{5}{c}{Interaction} 
& \multicolumn{4}{c}{GT Difference}  \\  
\cmidrule(lr){3-7} \cmidrule(lr){8-11}  
Sampling Method & Conditioning Method
& $C_{prec}\uparrow$ & $C_{rec}\uparrow$ & $C_{F_1}\uparrow$ & $C_{\%}$ & $P_{hand}\downarrow$ 
& MPJPE$\downarrow$  & $T_{root}\downarrow$  & $T_{obj}\downarrow$ & $O_{obj}\downarrow$ \\ 
\midrule

KNN & Embedding Fusion       & 0.79 & 0.67 & 0.69 & 0.58 & 0.61 & 14.79 & 20.15 & 11.52 & 1.07 \\
Uniform & Cross-Attention     & 0.19 & 0.22 & 0.20 & 0.23 & 0.59 & 29.89 & 84.56 & 110.77 & 1.23 \\
KNN & Cross-Attention        & \textbf{0.82} & \textbf{0.75} & \textbf{0.77} & \textbf{0.64} & 0.66 & \textbf{10.64} & \textbf{11.62} & \textbf{6.39} & \textbf{0.81} \\
\bottomrule 
\end{tabular}
} 
}
\end{table*}

\subsection{Discussion}

To further validate the effectiveness of our pipeline, we evaluate a variant (GenHOI-GT) where we omit the 3D HOI keyframe recovery stage during inference and instead use ground-truth 3D HOI keyframes from the dataset as the conditional input to ContactDM. The corresponding results are reported in Table~\ref{tab:single_window_cmp_unseen}. Despite relying on predicted keyframes in the full GenHOI model, it achieves performance comparable to the GenHOI-GT variant. This not only highlights the robustness of our ContactDM but also demonstrates the reliability of the 3D HOI keyframes generated in the first stage.

\section{Conclusion and Limitation}
In summary, we propose a novel two-stage framework for text-driven 4D HOI synthesis on unseen objects. By decomposing the task into (1) recovering 3D HOI keyframes for unseen objects and (2) interpolating sparse keyframes into temporally coherent 4D HOI sequences, our approach effectively reduces reliance on large-scale 4D HOI datasets. During temporal modeling, our contact-aware diffusion model explicitly leverages interaction priors during 4D generation, ensuring robust generalization to diverse object geometries.

Although GenHOI generates realistic 4D HOI sequences, it still has some limitations. The GenHOI framework is less adept at depicting multi-object interactions. While it excels at generating realistic and semantically consistent 4D human-object interactions for single-object scenarios, it struggles with more intricate interaction sequences. Additionally, it raises the risk of encoding incorrect contact information, as the model may struggle to distinguish relevant and irrelevant contact points in cluttered scenes.

 \bibliographystyle{plain} 
 \bibliography{main}


\appendix 
\newpage
\section{Technical Appendices and Supplementary Material}

\label{appendix}
In this section, we categorize our discussion into four main parts.
\subsection{Evaluation Metrics}
We employ Human Motion Metrics, Interaction Metrics, and GT Difference Metrics to comprehensively evaluate the quality of our generated 4D Human-Object Interaction (HOI) sequences.

\noindent\textbf{Human Motion Metric:} This metric integrates three key components: the Foot Sliding Score (FS), Fréchet Inception Distance (FID), and R-precision (Rprec). The FS quantifies foot sliding artifacts by computing the weighted average of accumulated translation in the xy-plane, following the methodology of \cite{he2022nemf}, and is measured in centimeters. The FID and Rprec are adopted from prior work \cite{guo2022generating}, where FID evaluates the distributional discrepancy between generated and real motions, and $R_{\text{prec}}$ assesses the alignment between textual descriptions and generated motions through a retrieval-based approach. Together, these metrics provide a comprehensive evaluation of motion quality, realism, and semantic consistency.

\noindent \textbf{Interaction Metric:} For contact accuracy, the metrics include precision ($C_{\text{prec}}$), recall ($C_{\text{rec}}$), and F1 score ($C_{F_1}$), which evaluate the correctness and completeness of detected contacts. Additionally, the contact percentage ($C\%$) measures the proportion of frames where contact is detected. The penetration score ($P_{\text{hand}}$) quantifies the extent of hand-object penetration by averaging the negative distance values (representing penetration) from the object’s Signed Distance Field (SDF), measured in centimeters.

\noindent \textbf{GT Difference Metric:} This metric is a metric category which quantifies the deviation of generated results from ground truth motion. It comprises the mean per-joint position error (MPJPE), the translation error of the root joint (\(T_{\text{root}}\)), and the object position error (\(T_{\text{obj}}\)), all computed as the Euclidean distance between predicted and ground truth positions in centimeters (cm). Additionally, it includes the object orientation error (\(O_{\text{obj}}\)), following prior works \cite{li2024controllable}.

\subsection{Object-AnchorNet}
\noindent \textbf{Datasets:} We construct a hybrid dataset by unifying \textit{BEHAVE}\cite{bhatnagar2022behave}, \textit{GRAB}\cite{taheri2020grab}, and \textit{Open3DHOI}\cite{wen2025reconstructing}, comprising:
\begin{itemize}
\item 16 human subjects—8 from \textit{GRAB}, 7 from the official training split of \textit{BEHAVE}, and 1 aggregated subject from \textit{Open3DHOI}, where all humans are reconstructed using a consistent 3D pipeline with identical parametric templates;
\item 173 object classes—20 from \textit{BEHAVE}, 20 selected from \textit{GRAB}, and 133 from \textit{Open3DHOI}.
\end{itemize}
Following \cite{petrov2023object}, we downsample \textit{GRAB} and \textit{BEHAVE} sequences to 10 FPS to extract single-frame HOI poses.

\noindent \textbf{Model Architecture:} Object-AnchorNet adapts the Object Pop-up \cite{petrov2023object}framework for human-centric 3D object inference, removing class-dependent components to focus on geometric encoding. The model processes a human point cloud through three key stages: (1) PointNet++-based object center prediction, (2) local neighborhood feature extraction via KNN, and (3) point-wise offset prediction for pose refinement, followed by Procrustes alignment for rigid transformation. This decomposition enables robust object localization from human interactions alone.
\begin{table}[h]
\caption{Object-AnchorNet Architecture}
\label{tab:architecture}
\centering
\begin{tabular}{ll}
\toprule
\textbf{Component} & \textbf{Function} \\
\midrule
Center Prediction & PointNet++ predicts object center $\mathbf{o}_P$ from input point cloud \\
Local Feature Extraction & KNN selects 3000 nearest points $\mathbf{P}_L$ for contact context \\
Offset Prediction & MLP predicts per-point shifts $\mathbf{S}_K$ for pose refinement \\
Procrustes Alignment & Computes rigid transformation $\mathbf{R},\mathbf{t}$ from deformed keypoints \\
\bottomrule
\end{tabular}
\end{table}


\newpage

\end{document}